\documentclass[sigconf]{acmart}

\acmYear{2026}\copyrightyear{2026}
\setcopyright{cc}
\setcctype[4.0]{by}
\acmConference[MMSys '26]{ACM Multimedia System Conference}{April 4--8, 2026}{Hong Kong, Hong Kong}
\acmBooktitle{ACM Multimedia System Conference (MMSys '26), April 4--8, 2026, Hong Kong, Hong Kong}
\acmDOI{10.1145/3793853.3795748}
\acmISBN{979-8-4007-2481-7/26/04}
\usepackage{booktabs}
\usepackage{amsmath,bm}
\usepackage{pifont}
\usepackage{balance}
\usepackage{threeparttable}
\usepackage{tikz}
\usepackage{subcaption}
\usepackage{multirow}
\usepackage[most]{tcolorbox}
\usepackage{mdframed}
\usepackage{graphicx}
\usepackage{xspace}
\usepackage{enumitem}
\usepackage{xcolor}
\usepackage{url}

\usepackage{breakurl}
\PassOptionsToPackage{hyphens}{url}\usepackage{hyperref}

\hypersetup{
    colorlinks=true,
    linkcolor=magenta,
    filecolor=magenta,      
    urlcolor=magenta,
    citecolor=magenta,
}
\setlist*{itemjoin={{, }}, itemjoin*={{, and }}}

\graphicspath{{assets/}{figures/}{images/}{pictures/}{./}}

\newcommand{\sysname}{\textsc{ARCADE}\xspace} 
\newcommand{\para}[1]{\noindent  {\bf #1}}

\newcommand{\cmark}{\small\ding{51}}   
\newcommand{\xmark}{\small\ding{55}}   
\newcommand{\pmark}{\small$\vartriangle$} 

\newcommand{\circlednumber}[1]{%
  \tikz[baseline=(char.base)]{
    \node[
      shape=circle,
      draw,
      fill=black,
      text=white,
      inner sep=1pt,
      minimum size=1em
    ] (char) {\textbf{#1}};}}

\title{AR as an Evaluation Playground: Bridging Metric and Visual Perception of Computer Vision Models}

\author{Ashkan Ganj}
\affiliation{%
  \institution{Worcester Polytechnic Institute}
  \city{Worcester}
  \state{MA}
  \country{USA}}
\email{aganj@wpi.edu}

\author{Yiqin Zhao}
\authornote{The work was done while at Worcester Polytechnic Institute.}
\affiliation{%
  \institution{Rochester Institute of Technology}
  \city{Rochester}
  \state{NY}
  \country{USA}}
\email{yzigm@rit.edu}

\author{Tian Guo}
\affiliation{%
  \institution{Worcester Polytechnic Institute}
  \city{Worcester}
  \state{MA}
  \country{USA}}
\email{tian@wpi.edu}

\begin{abstract}
Quantitative metrics are central to evaluating computer vision (CV) models, but they often fail to capture real-world performance due to protocol inconsistencies and ground truth noise. While visual perception studies can complement these metrics, they often require end-to-end systems that are time-consuming to implement and setups that are difficult to reproduce. We systematically summarize key challenges in evaluating CV models and present the design of \sysname\footnote{ \href{https://github.com/cake-lab/ARCADE}{\textcolor{magenta}{https://github.com/cake-lab/ARCADE}}}, an evaluation platform that leverages augmented reality (AR) to enable easy, reproducible, and human-centered CV evaluation.
\sysname uses a modular architecture that provides cross-platform data collection, pluggable model inference, and interactive AR tasks, supporting both metric and visual perception evaluation. We demonstrate \sysname through a user study with 15 participants and case studies on two representative CV tasks, depth and lighting estimation, showing \sysname can help reveal perceptual flaws in model quality that are often missed by traditional metrics. We also evaluate \sysname's usability and performance, showing its flexibility as a reliable real-time platform. 

\end{abstract}


\begin{teaserfigure}
  \centering
  \includegraphics[width=0.9\linewidth]{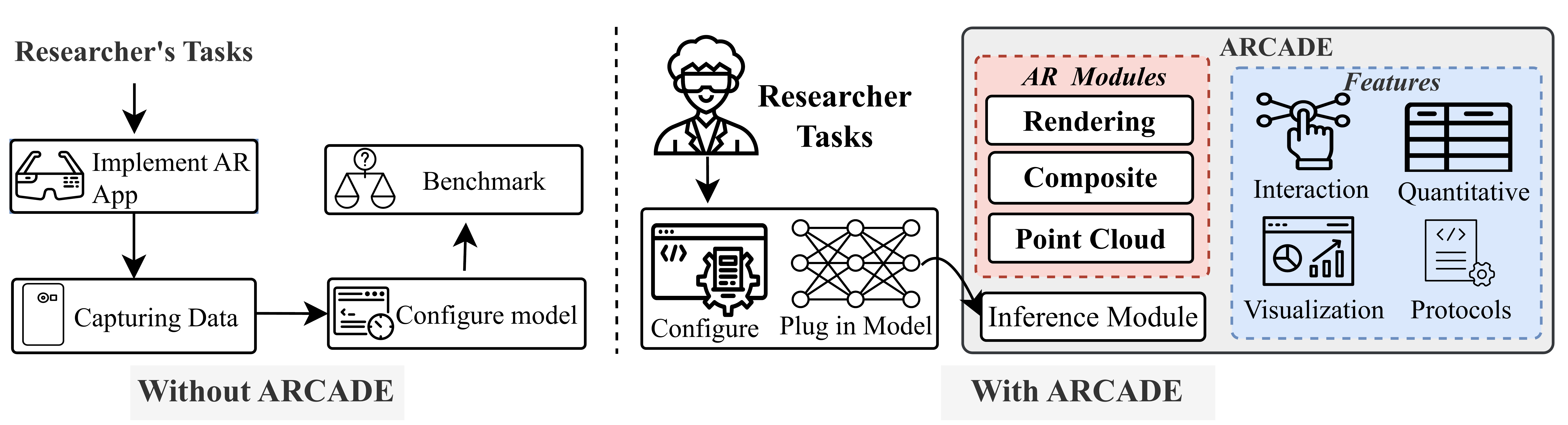} 
  \caption{CV model evaluation workflow comparisons.
  \textnormal{(Left): without \sysname, researchers need to build an AR app, collect data, configure models, and design benchmarks, which often leads to inconsistent, irreproducible assessments. (Right): With \sysname, researchers only need to configure experiment protocols and provide the CV models to conduct end-to-end metric and perception evaluations. 
  }
  }
  \label{fig:teaser}
\end{teaserfigure}

\begin{document}
\maketitle

\vspace{-0.2cm}
\section{Introduction}
\label{sec:introduction}

Artificial Intelligence(AI)~\cite{watson2023implict, 10.1145/3544548.3581500, nouraniboosjin2025uncertainty, xu2024multimodal, nouraniboosjin2024too} and more specifically, Computer vision(CV)~\cite{Ganj_2025_WACV, depth_anything_v2, depth_anything_v1, https://doi.org/10.48550/arxiv.2302.12288,  10.1115/MSEC2025-155144} models are central to applications across different fields such as AR~\cite{10.1145/3638550.3641122}, robotics~\cite{}, and autonomous systems. 
However, the evaluation methodology for testing these models in real-world settings is fragmented and often misleading.

First, standard data-driven benchmarks are sensitive to seemingly minor evaluation choices. For example, small, often undocumented decisions in data processing---such as resizing, masking, and clipping---can nonlinearly shift scores and even reorder model rankings (\S\ref{subsec:evaluation_limitations}). Differences in dataset splits further complicate cross-paper comparisons, often requiring costly retraining using a common protocol. 
In addition, benchmarks treat ground truth (GT) as definitive, yet GT quality is inherently limited by capture hardware and collection conditions. As we show in \S\ref{subsec:data_labels}, even minor changes in the environment (e.g., adjusting ambient lighting) can distort sensor GT and increase a model's reported errors by more than 30\%, making it hard to separate true model performance from sensor-induced bias.

Second, task-specific numerical metrics do not always translate to a model's performance in an end-to-end application context. As we demonstrate in \S\ref{sec:caseStudy}, models with superior benchmark scores can still exhibit pronounced perceptual failures when deployed, such as misplacing virtual objects or producing unstable geometry in AR applications or 3D reconstruction (Fig.~\ref{fig:depth_cs_pointcloud}). Closing this gap requires evaluation setups that more closely reflect the target experience, e.g., interactive, application-focused, and perception-centered, rather than relying solely on metrics. However, our user study indicates that such application-driven evaluation remains relatively rare in practice: only 47\% of participants reported usually testing models in real-world application scenarios, and engineering effort and lack of knowledge were the dominant barriers. Thus, the combination of unreliable quantitative metrics and the high barrier to application-level, perception-based testing discourages thorough evaluation, making it difficult to assess a CV model’s utility in practical contexts.

Motivated by the above challenges, we introduce \sysname, which is designed to make CV evaluation more reproducible, perception-centered, and plug-and-play. \sysname's design also reflects the needs reported in our user study (\S\ref{subsec:background_study}), where practitioners emphasized protocol consistency, visual inspection, and reduced engineering effort as key requirements. \sysname consists of four components: data capture, pluggable inference models, AR tasks, and visualization tools. It supports both existing datasets and live or recorded data via our cross-platform client, and exposes a pluggable inference interface (REST/Docker) for easy model integration and side-by-side visual comparison. Fig.~\ref{fig:teaser} shows the researcher workflow with and without \sysname.


We evaluate \sysname in three ways to test its practicality, usability, and effectiveness. First, we perform an end-to-end evaluation and show that \sysname can support real-time AR workflows. For example, on a university WiFi network with an AR device, edge server, and web client, the render-composite loop achieves average latencies of 5.2~ms at 640$\times$480 and about 20~ms at 1920$\times$1080, with interaction latencies between 7.5~ms and 18~ms.

Second, we conduct a user study with 15 CV/ML researchers to understand practical challenges they encounter in CV evaluation and how effective ARCADE’s design is in addressing them. Participants reported high overall satisfaction (4.20/5). They rated \sysname very effective for judging task-level suitability of depth models (4.67/5) and for discovering failure cases more quickly (4.53/5). Moreover, researchers reported \sysname can reduce the engineering effort required for real-world evaluation (4.33/5), and most features are easy to use and responsive.

Third, we use \sysname to perform case studies on depth and lighting estimation, two core components of AR rendering with distinct evaluation needs, where multiple state-of-the-art (SoTA) models are assessed using standard numerical metrics alongside interactive AR tasks. 
These case studies show that \sysname can effectively expose geometric errors such as depth discontinuities and misaligned occlusions, as well as visually implausible lighting, even when benchmark results appear strong.

In summary, we make the following key contributions:

\begin{itemize}[leftmargin=*,topsep=0pt,noitemsep]

\item \textbf{Empirical characterization of CV evaluation limitations.} We identify and characterize key challenges in common CV evaluation practices, including inconsistencies in dataset usage, underuse of perceptual feedback, and the high effort required for in-the-wild testing. Our user study echoes these limitations: respondents reported only moderate confidence that benchmark metrics transfer to real applications, tended to view metrics and leaderboard scores as insufficient, and strongly agreed that results are sensitive to protocol choices and sensing hardware.

\item \textbf{Design of \sysname, a modular and perception-centered evaluation framework.} 
\sysname addresses the evaluation challenges in the CV workflow identified by our empirical analysis and user study. \sysname provides an easy-to-use way for CV practitioners to evaluate model performance using custom, consistent protocols across datasets and captured data. Moreover, it lowers the barrier of perception evaluation by allowing interactive debugging via low-latency AR streaming and supporting perception analysis in the context of AR tasks.

\item \textbf{Evaluation insights and open-source artifacts.} Through case studies on depth and lighting estimation, we show how \sysname reveals perceptual discrepancies that standard metrics fail to capture. A complementary user study also supports the case study results and shows that \sysname is rated highly overall and is perceived as effective for judging task suitability and discovering failure cases more quickly, while reducing the engineering effort of real-world evaluation. We will open source \sysname and a dataset of over 2000 AR frames collected using our custom mobile setup (iPhone~14 Pro), including RGB, LiDAR depth, intrinsics, extrinsics, and virtual object metadata.

\end{itemize}

\section{Related Work}
\label{sec:related}

\begin{table}[t]
\centering
\caption{Feature comparison between \sysname and prior XR/AR-related evaluation frameworks.}
\label{tab:framework_compare}
\small 
\begin{tabular}{lcccccccc}
\toprule
 & \textbf{C/R}\! & \textbf{RT}\! & \textbf{PP}\! & \textbf{CMC}\! & \textbf{IT}\! & \textbf{MD}\! & \textbf{SM}\! & \textbf{OS}\! \\
\midrule
\textbf{\sysname} & \cmark & \cmark & \cmark & \cmark & \cmark & \cmark & \cmark & \cmark \\ \midrule
ILLIXR \cite{Huzaifa2021-jj}     & \xmark & \cmark & \cmark & \xmark & \xmark & \pmark & \cmark & \cmark \\
XRBench \cite{Kwon2022-vm-mlsys}    & \xmark & \xmark & \cmark & \cmark & \xmark & \xmark & \xmark & \cmark \\
ARGUS \cite{10305427}      & \cmark & \cmark & \cmark & \pmark & \cmark & \cmark & \cmark & \cmark \\
ExpAR \cite{Ganj2023-nt}      & \cmark & \cmark & \pmark & \pmark & \pmark & \cmark & \cmark & \xmark \\
CoMIC \cite{Han2022-bh}      & \pmark & \cmark & \pmark & \pmark & \xmark & \cmark & \pmark & \pmark \\
TOM \cite{janaka2024tom}        & \xmark & \cmark & \cmark & \xmark & \cmark & \pmark & \pmark & \pmark \\
\bottomrule
\end{tabular} \\[2pt]
\begin{minipage}{\linewidth}
\small
\raggedright
C/R: capture \& replay; RT: real-time streaming; PP: plug-and-play modules; CMC: cross-model comparison; IT: interactive AR tasks; MD: multi-device clients; SM: multi-sensor modalities; OS: open source. 
\cmark: fully supported; \pmark: partially / limited; \xmark: not supported.
\end{minipage}
\vspace{-2.5em}
\end{table}

There are several parallel efforts to improve the numerical evaluation of CV models. Zhang et al. propose LPIPS, a learned perceptual metric that better correlates with human judgments~\cite{zhang2018unreasonableeffectivenessdeepfeatures}. Giroux et al.~\cite{giroux2024towards} develop a psychophysical framework for evaluating lighting estimation, showing that standard image-quality metrics often fail to match human perception. Standardized benchmarks can also reduce evaluation drift: for example, COCO’s official metrics~\cite{coco_eval} are widely adopted for object detection, and RobustBench~\cite{croce2021robustbench} addresses pitfalls in adversarial robustness evaluation.

Beyond fair numerical evaluation, it is equally important to understand how these values translate to human-perceived performance. Only a few works explicitly connect metrics to perception. HuPerFlow~\cite{yang2025huperflow} aligns human-perceived motion with optical flow performance, and HYPE~\cite{Zhou2019-hd} evaluates generative realism via large-scale crowdsourcing. These efforts highlight the value of perception-centered evaluation, but are tied to specific tasks and do not provide a general, application-grounded workflow.

As CV models are ultimately deployed in applications, understanding in-the-wild behavior is also crucial. Out-of-distribution and zero-shot testing are increasingly used to probe generalization, but collecting high-quality OOD datasets is difficult~\cite{mukhoti2023raising}, and zero-shot performance on new benchmarks~\cite{Rafi2022-mc,depth_anything_v2,Ganj_2025_WACV} does not always predict how models behave in real AR scenes. Systems like TOM~\cite{janaka2024tom} and ARGUS~\cite{10305427} help prototype CV-based wearable AR assistants, making it easier to study model behavior in specific application contexts. ARFlow~\cite{zhao2024arflow} and ExpAR~\cite{Ganj2023-nt} simplify capture and visualization of high-quality sensor data for interactive debugging. XR-focused evaluation frameworks such as ILLIXR~\cite{Huzaifa2021-jj}, CoMIC~\cite{Han2022-bh}, and XRBench~\cite{Kwon2022-vm-mlsys} make it easier to test numerical accuracy and latency in XR pipelines, but provide limited support for cross-model visual comparison and perception-based analysis.

\sysname is designed to address these limitations by combining capture-and-replay, built-in metrics, and custom experiment protocols with interactive AR tasks and live streaming. It allows researchers to (i) run fair, reproducible comparisons across models and devices using shared data and standardized preprocessing, and (ii) directly inspect model behavior through AR tasks such as object placement, occlusion, and point-cloud visualization. As summarized in Table~\ref{tab:framework_compare}, \sysname uniquely provides capture/replay, real-time streaming, plug-and-play modules, cross-model comparison, interactive AR tasks.

\begin{table}[t]
\centering
\caption{Protocol sensitivity on NYU Depth V2\cite{Silberman:ECCV12}. 
\textnormal{Baseline is \emph{Clip \& Mask} (clipping depth range + removing black borders). Values in parentheses show the change from the baseline.}}
\label{tab:prot_sensitivity_both}
\small
\begin{tabular}{@{}llccc@{}}
\toprule
\textbf{Model} & \textbf{Metric} 
& \textbf{Baseline} 
& \textbf{No depth clip} 
& \textbf{No black mask} \\
\midrule
\multirow{3}{*}{ZoeDepth} 
& RMSE $\downarrow$   & \textbf{0.270} & 0.275 (+0.005)   & --- \\
& AbsRel $\downarrow$ & \textbf{0.075} & 0.075 (+0.000)   & --- \\
& $\delta_1$ $\uparrow$ & \textbf{0.955} & 0.954 (\,-0.001) & --- \\
\midrule
\multirow{3}{*}{HybridDepth} 
& RMSE $\downarrow$   & \textbf{0.128} & 0.132 (+0.004)   & 0.129 (+0.001) \\
& AbsRel $\downarrow$ & \textbf{0.039} & 0.040 (+0.001)   & 0.041 (+0.002) \\
& $\delta_1$ $\uparrow$ & \textbf{0.995} & 0.995 (+0.000)   & 0.996 (+0.001) \\
\bottomrule
\end{tabular}
\vspace{-0.5cm}
\end{table}

\section{Challenges in Evaluating CV models}
\label{sec:challenges}

Task-specific metrics are the dominant way to report and compare the performance of computer vision models, but there is little consensus on which metrics to use or how to compute them~\cite{piccinelli2024unidepth}. For example, one depth-estimation paper focuses on AbsRel and $\delta{<}1.25$~\cite{lu2025align3r}, whereas another emphasizes RMSE and AbsRel but omits $\delta$-threshold accuracies~\cite{yan2025synthetic2real}, complicating fair comparison and making accuracy gains hard to interpret. In the rest of this section, we highlight two main problems. First, we show that data-driven benchmarks are highly sensitive to hidden protocol choices, dataset splits, alignment strategies, and sensor-dependent ground truth, which makes scores brittle and reproducibility difficult (\S\ref{subsec:evaluation_limitations}). Second, even when these pipelines are carefully controlled, numerical metrics often fail to reflect how humans perceive model behavior in real applications, leading to a persistent gap between metric and perception (\S\ref{sec:perceptionGap}).

\begin{table}[t]
\centering
\caption{Performance comparison on the NYU Depth V2~\cite{Silberman:ECCV12}}
\label{tab:performance_nyu_SIDE}
\resizebox{\columnwidth}{!}{
\begin{tabular}{@{}l|lrrrrr@{}}
\toprule
\textbf{Model}  & \textbf{RMSE} $\downarrow$ & \textbf{AbsRel} $\downarrow$ & \textbf{$\delta_1$} $\uparrow$ & \textbf{$\delta_2$} $\uparrow$ & \textbf{$\delta_3$} $\uparrow$ \\
\midrule
DPT~\cite{Ranftl2021}  & 0.357 & 0.104 & 0.904 & 0.988 & 0.998 \\
ZoeDepth~\cite{https://doi.org/10.48550/arxiv.2302.12288}\tnote{$\dagger$}  & 0.270 & 0.075 & 0.960 & 0.995 & 0.999 \\
Depth AnythingV1~\cite{depth_anything_v1} & {0.206} & {0.056} & {0.980} & {0.998} & {0.999} \\
Depth AnythingV2~\cite{depth_anything_v2} & \textbf{{0.206}} & \textbf{{0.056}} & \textbf{{0.984}} & \textbf{{0.998}} &\textbf{ {1.000}} \\

\bottomrule
\end{tabular}
}
\vspace{-0.5cm}
\end{table}

\begin{figure}[t]
    \centering
    \includegraphics[width=\linewidth]{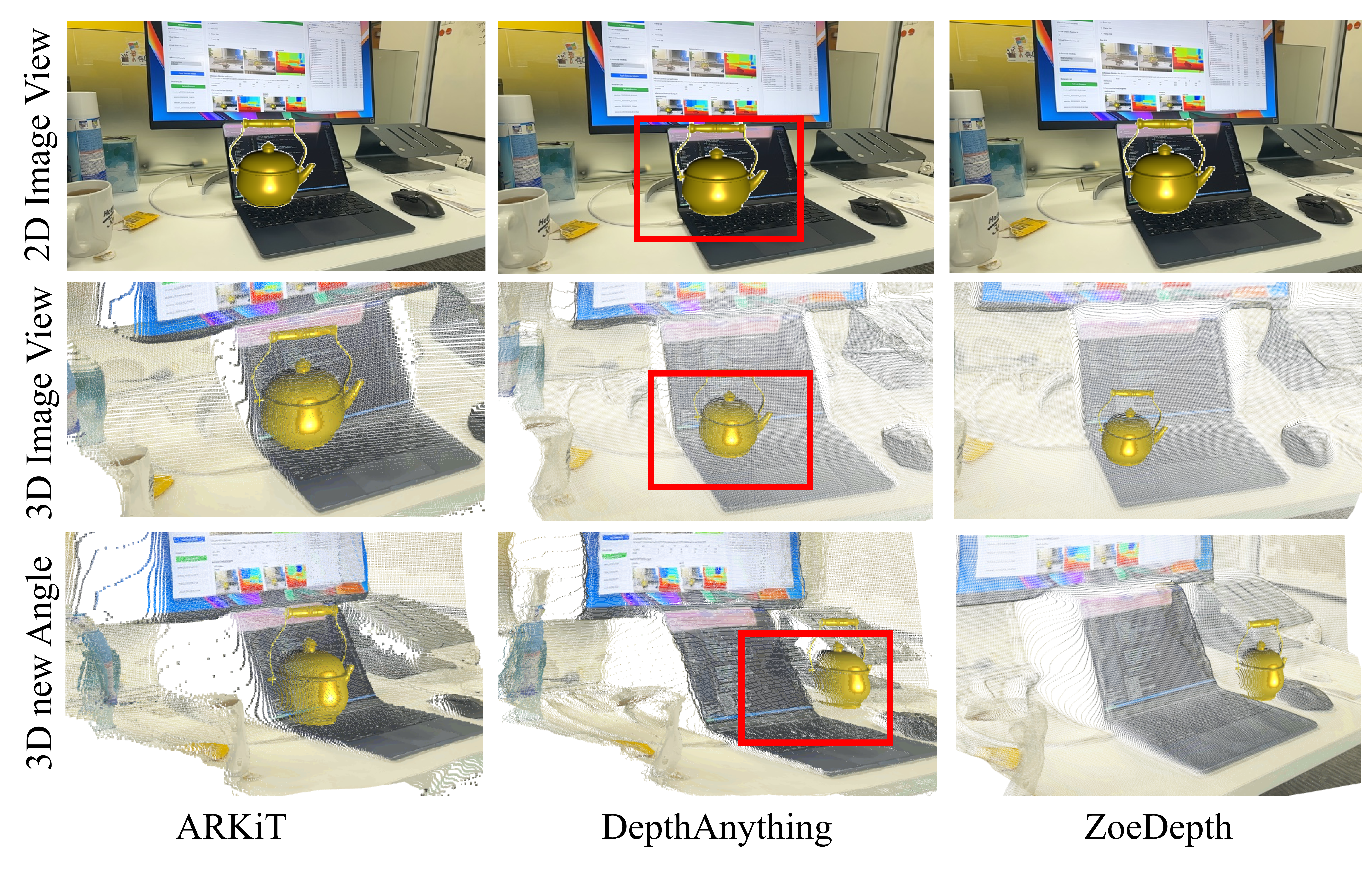}
    \caption{Viewing teapot placement from different angles by using \sysname's point cloud features. 
    }
    \label{fig:depth_cs_pointcloud}
    \vspace{-0.5cm}
\end{figure}

\subsection{Limitations of Data-Driven Benchmarking}
\label{subsec:evaluation_limitations}
A model’s reported score often depends as much on hidden evaluation choices as on the model itself. Small variations in resizing, normalization, masking, tone mapping, or clipping shift metrics and can even flip rankings, making apples-to-apples comparisons difficult and replication hard and time-consuming.

\subsubsection{Hidden choices in processing}
Performance metrics are highly sensitive to hidden choices made during data preprocessing and postprocessing, e.g., image resizing, input normalization, and validity masks. These decisions can systematically tilt final scores and even reorder model rankings. As shown in Table~\ref{tab:prot_sensitivity_both} for depth estimation, simply disabling depth-range clipping for \textit{ZoeDepth}~\cite{https://doi.org/10.48550/arxiv.2302.12288} and \textit{HybridDepth}~\cite{Ganj_2025_WACV} shifts RMSE by about \(2\%\) because the valid-pixel mask changes. In a second check, leaving sensor-specific black borders unmasked for \textit{HybridDepth} increases AbsRel from \textbf{0.039} to \textbf{0.041}. These small numerical shifts can swap the ordering between close competitors, undermining apples-to-apples comparisons when pipelines are not fixed and transparent.
Similarly, in lighting estimation, hidden choices often skew evaluation results on rendered images. For example, the selection of the gamma correction ratio during HDR tone mapping has non-negligible impacts on rendering. 
In an experiment in which we tested the DiffusionLight~\cite{Phongthawee2023DiffusionLight} model's impact on mirror object rendering, 
we observed that setting $\gamma=2.2$ (default is $\gamma=2.4$) reduces the RMSE from \textnormal{0.43} to \textnormal{0.41}, whereas setting $\gamma=2.6$ increases the RMSE to \textnormal{0.45}.

\subsubsection{Dataset-specific choices}
Reproducibility is further complicated by inconsistent data splits. For example, for the NYU Depth V2 dataset~\cite{Silberman:ECCV12}, some papers~\cite{maximov2020focus} use the standard split, which contains 1449 pairs of aligned RGB and depth frames. On the other hand, papers such as HybridDepth, ZoeDepth, and DepthAnything use a different data split followed by the BTS paper~\cite{lee2019big}, which contains 654 images for testing and 24k images for training. Due to this, results from the DefocusNet paper and ZoeDepth are not directly comparable. The only way to directly compare the models is to retrain them all with a specific split, which can be very costly.

\begin{figure}[t]
  \centering
  \begin{subfigure}[b]{0.47\columnwidth}
    \centering
    \includegraphics[width=\linewidth]{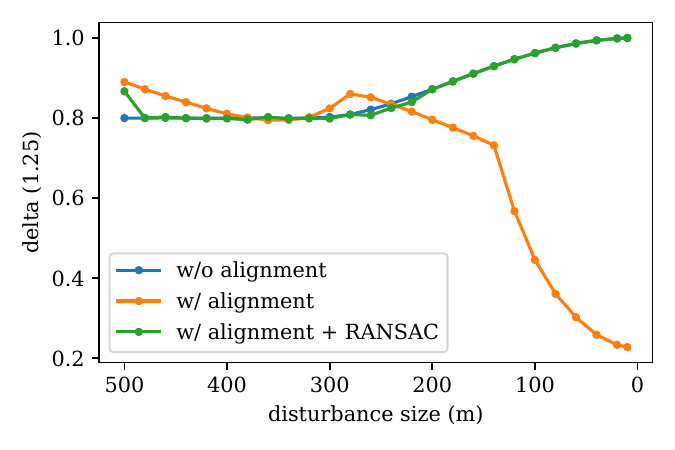}
    \caption{Alignment disagreement~\cite{li2025benchdepth}}
    \label{fig:delta_vs_sigma}
  \end{subfigure}
  \begin{subfigure}[b]{0.47\columnwidth}
    \centering
    \includegraphics[width=\linewidth]{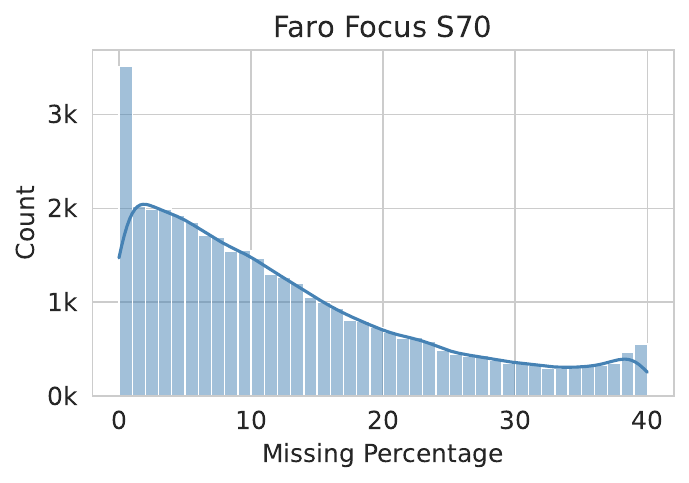}
    \caption{Missing GT depth}
    \label{fig:missing_arkitscenes}
  \end{subfigure}

  \caption{(a) Outliers substantially change alignment and monotonicity. (b) A histogram of missing depth percentage in each sample from the ARKitScenes dataset.}
  \label{fig:alignment_and_missing}
\vspace{-0.5cm}
  
\end{figure}

\subsubsection{Alignment in Different Spaces}

Depth models produce outputs in different spaces: metric depth (absolute units), relative depth (unknown global scale and shift), or disparity. To enable direct comparison with a single ground truth, researchers commonly apply a global scale-and-shift alignment, typically solved with least-squares (LS) regression. However, as noted in BenchDepth~\cite{li2025benchdepth}, 
global LS scale-and-shift alignment assumes a simple linear relationship between predicted and ground-truth depth, whereas inverse-depth and disparity are inherently non-linear, which can cause problems in model comparison.
Fig.~\ref{fig:delta_vs_sigma} shows the sensitivity of the $\delta(1.25)$ metric following the BenchDepth setup~\cite{li2025benchdepth}. As the disturbed region becomes smaller (fewer but stronger outliers), the unaligned score improves (fewer pixels are wrong), while the LS-aligned score worsens as the fitted global scale and shift are pulled toward the outliers. 
RANSAC+LS mitigates this effect by ignoring many outliers during alignment, but does not eliminate the distortion entirely. 
Consequently, alignment can alter or even reverse metric trends, making cross-space model comparisons unreliable. 

\subsubsection{Data \& Labels}
\label{subsec:data_labels}

Benchmarks treat ground truth (GT) as definitive, but its quality is often limited by capture hardware and collection conditions. Depth sensors, for example, routinely fail on transparent and low-albedo surfaces, producing GT with holes and biased measurements; a model that correctly infers more complete geometry may then be penalized for disagreeing with this flawed data. This affects even high-end devices: the LiDAR in ARKitScenes~\cite{arkitscenes} exhibits substantial missing depth (Fig.~\ref{fig:missing_arkitscenes}).

\begin{figure}[t]
  \centering
  \includegraphics[width=\columnwidth]{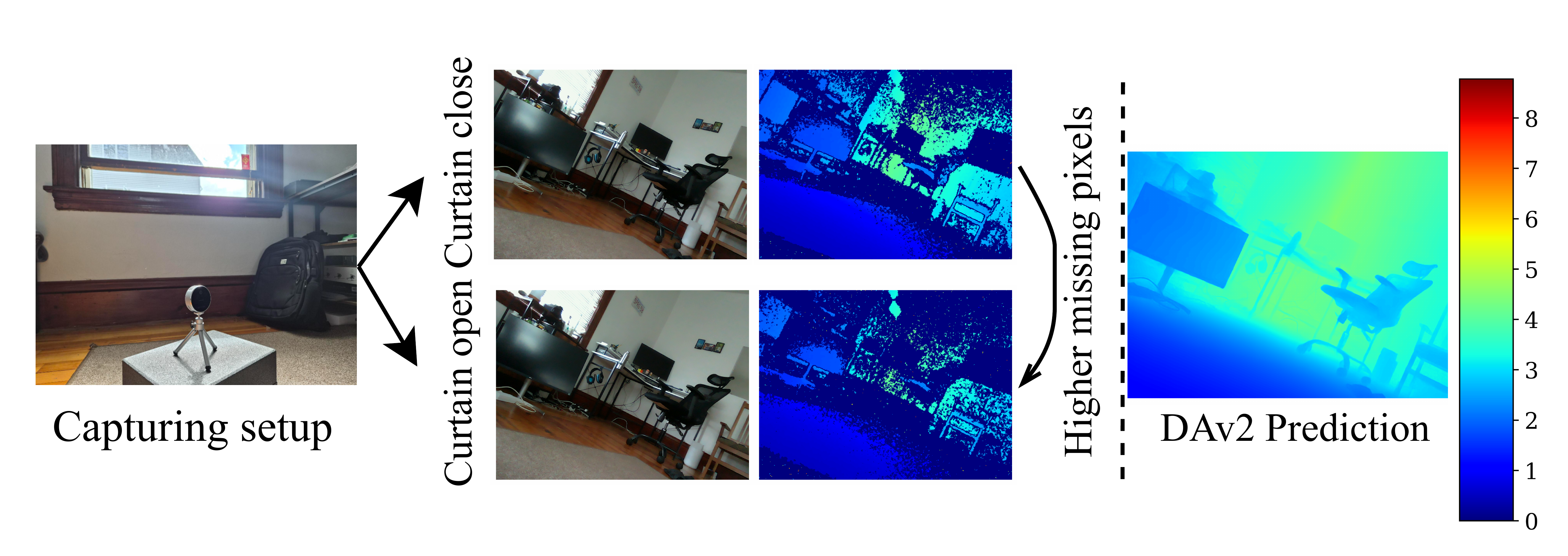}
  \caption{Ambient lighting sensitivity in evaluation with RealSense L515 ground truth and DepthAnythingV2 model. 
  \textnormal{
  Opening curtains increases the error for metrics: RMSE $+11.1\%$, MAE $+14.5\%$, AbsRel $+40.3\%$.}
  }
  \label{fig:lighting_bias_singlecol_errors}
\vspace{-0.5cm}
  
\end{figure}

Environmental factors further destabilize GT. Datasets collected over time cannot keep lighting and other conditions fixed. In Fig.~\ref{fig:lighting_bias_singlecol_errors}, we show an Intel RealSense L515 experiment where the camera and scene are fixed but ambient light is changed simply by opening or closing curtains. This small change degrades the sensor’s GT and increases DepthAnything’s errors (RMSE +11.1\%, MAE +14.5\%, AbsRel +40.3\%), showing how easily the sensor can bias evaluation and undermine benchmark reliability.

\begin{table}[t]
\centering
\caption{
Quantitative evaluation of environment maps and the three-sphere protocol. 
\textnormal{We test three representative lighting estimation models: XiheNet~\cite{xihe_mobisys2021}, StyleLight~\cite{wang2022stylelight}, and DiffusionLight~\cite{Phongthawee2023DiffusionLight} under the standard test setting with a single 75$^\circ$ horizontal FoV input. We noticed major misalignment on different evaluation metrics. Specifically, models that perform well on RMSE and Si-RMSE may not yield good angular error numbers.
}
}
\label{tab:eval_three_sphere_reorganized}
\resizebox{\columnwidth}{!}{
\begin{tabular}{llccc}
\toprule
\textbf{Target} & \textbf{Model} & \textbf{Angular Error $\downarrow$} & \textbf{Si-RMSE $\downarrow$} & \textbf{RMSE $\downarrow$} \\
\midrule
\multirow{3}{*}{\shortstack[l]{Env.\\Map}} 
    & XiheNet~\cite{xihe_mobisys2021} & 7.39 & 0.23 & 0.34 \\
    & StyleLight~\cite{wang2022stylelight} & 5.81 & 0.27 & 0.25 \\
    & DiffusionLight~\cite{Phongthawee2023DiffusionLight} & 6.76 & 0.26 & 0.20 \\
\toprule
\multirow{3}{*}{\shortstack[l]{Diffuse\\Sphere}} 
    & XiheNet~\cite{xihe_mobisys2021} & 11.8 & 0.12 & 0.21 \\
    & StyleLight~\cite{wang2022stylelight}        & 4.24 & 0.13 & 0.23 \\
    & DiffusionLight~\cite{Phongthawee2023DiffusionLight} & 2.14 & 0.14 & 0.20 \\
\midrule
\multirow{3}{*}{\shortstack[l]{Matte\\Sphere}} 
    & XiheNet~\cite{xihe_mobisys2021} & 12.2 & 0.17 & 0.23 \\
    & StyleLight~\cite{wang2022stylelight}        & 4.74 & 0.31 & 0.40 \\
    & DiffusionLight~\cite{Phongthawee2023DiffusionLight} & 3.42 & 0.33 & 0.36 \\
\midrule
\multirow{3}{*}{\shortstack[l]{Mirror\\Sphere}} 
    & XiheNet~\cite{xihe_mobisys2021} & 13.4 & 0.27 & 0.31 \\
    & StyleLight~\cite{wang2022stylelight}        & 6.78 & 0.55 & 0.51 \\
    & DiffusionLight~\cite{Phongthawee2023DiffusionLight} & 5.94 & 0.60 & 0.43 \\
\bottomrule
\end{tabular}}
\vspace{-2em}
\end{table}

\begin{figure*}[t]
    \centering
    \includegraphics[width=0.9\linewidth]{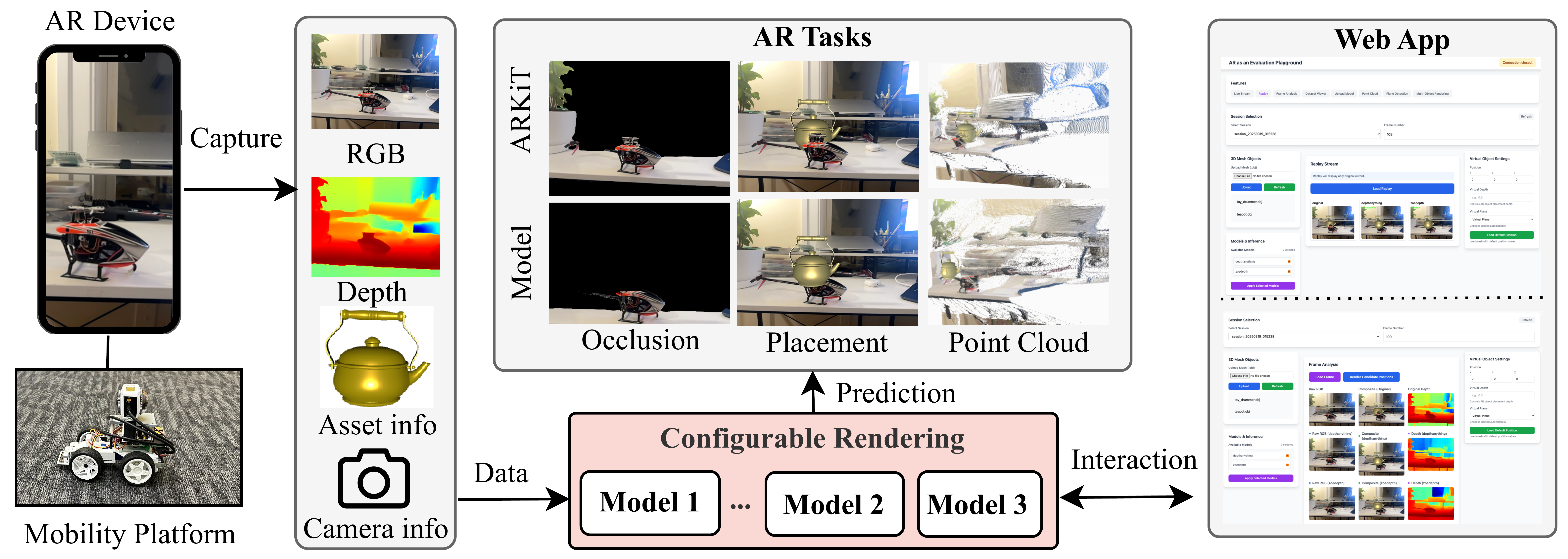}
\caption{
A simplified workflow of \sysname. 
\textnormal{
A scene is captured once, streamed to the configurable AR‑task engine, and rendered with two example depth models, ARKit and DepthAnythingV2 (DAv2). Researchers can visually inspect and interact with the AR tasks to iteratively design experiment protocols and perform perception evaluations.
}
}

    \label{fig:workflow}
    \vspace{-0.5cm}
\end{figure*}

\subsection{The Metric-Perception Gap}
\label{sec:perceptionGap}

While \S\ref{subsec:evaluation_limitations} shows that benchmark pipelines themselves are fragile, even carefully controlled scores do not necessarily translate into a good user experience. Our depth and lighting case studies illustrate this metric-perception gap: models that appear strong on standard metrics can still produce unstable geometry or visually implausible lighting estimation.

\para{Discrepancies in Depth Estimation.}
Table~\ref{tab:performance_nyu_SIDE} reports results on NYU~Depth~V2~\cite{Silberman:ECCV12} for recent state-of-the-art models such as DepthAnything~V2~\cite{depth_anything_v2} and ZoeDepth~\cite{https://doi.org/10.48550/arxiv.2302.12288}. Despite strong benchmark scores, these models exhibit pronounced perceptual failures when used in interactive AR tasks. As Fig.~\ref{fig:depth_cs_pointcloud} illustrates, models with superior RMSE can still produce visible artifacts: virtual objects are misplaced (appearing to float rather than rest on surfaces), scale is inconsistently estimated (we observe systematic metric overestimation that breaks spatial coherence), and object boundaries are blurred--see \S\ref{sec:caseStudy}, producing perceptually important occlusion errors that global metrics like RMSE tend to ignore. Notably, DepthAnything~V2, which is about 25\% better than ZoeDepth in RMSE, still misplaces objects in many scenes. For example, in the last row of Fig.~\ref{fig:depth_cs_pointcloud}, both models overestimate depth, causing a teapot (0.65m) to appear unnaturally close to the camera.

\para{Discrepancies in Lighting Estimation.}
Similarly, in our lighting estimation case study, we find that numerical metrics can be misleading or even contradictory. Different metrics often yield conflicting conclusions; for instance, Table~\ref{tab:eval_three_sphere_reorganized} shows quantitative results under the three-sphere protocol. Our evaluation reveals that different numerical metrics often highlight different aspects of lighting estimation performance. For instance, XiheNet~\cite{xihe_mobisys2021} achieves the lowest Si-RMSE in environment map-wise comparisons, yet it performs the worst in terms of angular error. This illustrates a key limitation of single-metric evaluation: while they provide convenient summaries, they are not expressive enough to capture the full spectrum of model behavior. And this is particularly true for visual realism. We suspect the reason is that metrics may emphasize different aspects, such as directional alignment or intensity differences, but none alone can fully account for how estimation errors manifest in rendered AR scenes. As a result, relying solely on these metrics can often lead to misleading or incorrect conclusions about a model’s true perceptual quality.

These results show that quantitative scores alone fail to predict user experience. This gap arises from inconsistent protocols, imperfect ground truth, and environment-dependent bias, motivating a simple evaluation approach that complements benchmarks with direct visual and interactive assessment.
\vspace{-0.2cm}
\section{\sysname Design}
\label{sec:design}

The challenges in Section~\ref{sec:challenges} show three concrete requirements for a better evaluation tool: reproducibility, perception-centered evaluation, and plug-and-play extensibility.  To achieve reproducibility, \sysname adopts a \emph{capture-once, evaluate-many} workflow: sensor data from a physical scene is captured once, with all required data and parameters, and then reused for any combination of models, tasks, and metrics, removing variability from repeated captures. To support perception-centered evaluation, \sysname embeds models in a set of reusable AR tasks so researchers can inspect artifacts under real-world scenarios. To keep engineering overhead low, \sysname provides pluggable interfaces making it straightforward to add new models, sensors, datasets, or AR tasks so the system can evolve without rewriting core components.

To illustrate these challenges, consider a CV researcher who wants to test a new depth model that performs well on a benchmark like NYU Depth V2~\cite{Silberman:ECCV12}. Without \sysname, they must build a custom evaluation app to assess its real-world impact on a task like AR occlusion rendering.
However, two key issues arise: First, the researcher might design and implement the AR application in their own way, which makes it difficult to compare any results that are published and tested with this workflow. Second, the researcher can't easily compare the occlusion rendering effects produced by different depth models in a controlled manner, e.g., using the same physical scene.

\sysname replaces this with a shared workflow (Fig.~\ref{fig:workflow}). The researcher uses our data-collection client to capture RGB, depth, and pose for real scenes; these recordings are fed into a configurable AR-task engine that runs one or more depth or lighting models through a single preprocessing stack and renders the corresponding AR tasks. The same captured session can be replayed with different models, viewed live through the web portal for interactive debugging, and stored for later use in perception studies under custom experiment protocols. In this way, \sysname makes the design requirements above concrete: it fixes the evaluation pipeline, embeds models in reusable AR tasks for perception-centered analysis, and provides plug-and-play hooks to add new models and tasks with minimal engineering effort.

\begin{figure*}[t]
  \centering
  \begin{subfigure}[b]{\columnwidth}
    \centering
    \includegraphics[width=\linewidth]{assets/depth_CS_screenshot.pdf}
    \caption{}
    \label{fig:depth_casestudy_screenshot}
  \end{subfigure}\hspace{0.01\columnwidth}
  \begin{subfigure}[b]{\columnwidth}
    \centering
    \includegraphics[width=0.9\linewidth]{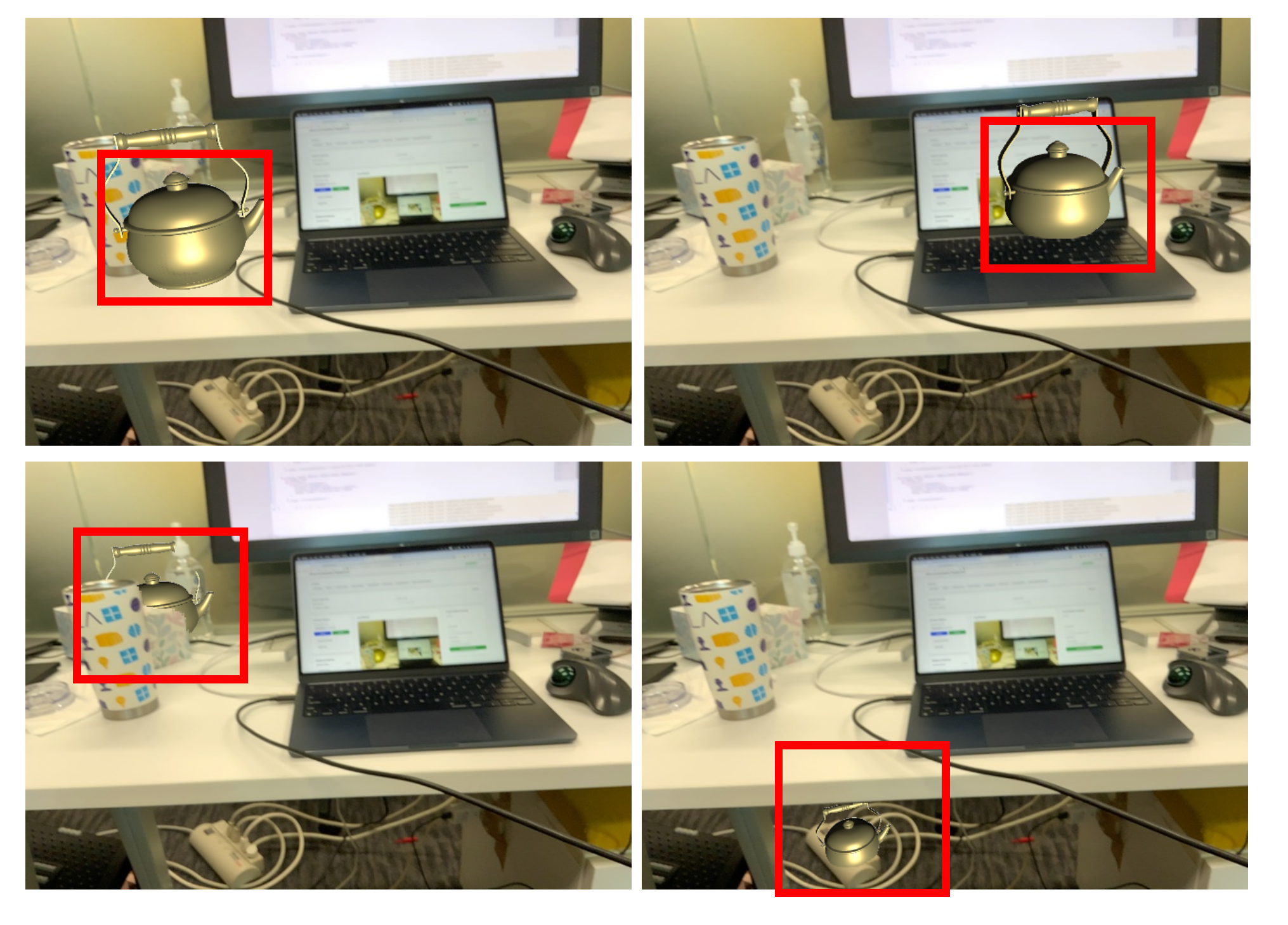}
    \caption{}
    \label{fig:autoPlacement}
  \end{subfigure}
  \caption{
  Visualization of \sysname's features.
  \textnormal{(a) Screenshot of \sysname's UI showing metrics alongside depth and object placement visualizations for a captured scene. (b) Illustration of the automatic virtual object re-rendering pipeline. We detect valid placement planes and place assets only on valid surfaces without re-capturing the scene on the client.}
  }
  \label{fig:depth_cs_pointcloud}
\vspace{-0.5cm}
  
\end{figure*}

\vspace{-0.2cm}
\subsection{Pluggable CV Model Inference}
\label{subsec:PlugInference}

A central goal of \sysname is to allow researchers to test and compare CV models \emph{within real-world AR contexts}. To support this, we design a pluggable inference module that is compatible with third-party model-hosting ecosystems (e.g., Hugging Face) as well as containerized Docker deployments that bundle custom environments and dependencies. This design reduces the overhead of setting up real-world evaluations and enforces consistent evaluation protocols, addressing the reproducibility and comparability challenges discussed in \S\ref{subsec:evaluation_limitations}.
Currently, \sysname provides built-in integration for two types of CV models: depth estimation and lighting estimation. We focus on these tasks because they are essential to many AR applications, which makes them good candidates to evaluate in the context of AR. To plug their depth or lighting estimation models into \sysname's AR task, researchers need to provide the model file and also customize the PyTorch-based template to implement their own \texttt{infer} function, which specifies how to perform model loading, pre-processing, and inference, or with a Dockerized container.
\vspace{-0.4mm}

\subsection{Interactive AR Task Support}
\label{subsec:ARTask}

To bridge the gap between benchmark performance and real-world usefulness and address \S\ref{sec:perceptionGap}, \sysname goes beyond traditional dataset-based metrics and introduces \emph{application-driven, perception-centered evaluation}. Specifically, we implement three interactive AR tasks that allow researchers to test their CV models, such as depth or lighting estimation, within realistic rendering contexts. These tasks are designed to expose perceptual failures that standard metrics may miss, such as object misplacement, incorrect occlusions, or lighting artifacts. As shown in the center of Fig.~\ref{fig:workflow}, researchers can use their model in multiple AR tasks to evaluate performance in settings that reflect real AR use cases.

\para{Object Rendering.} The first task is virtual object placement and rendering using the model prediction. Object placement is a common feature in many AR applications, like furniture shopping, and often requires the cooperation of multiple types of CV models~\cite{Yi2020-na}. This task helps researchers evaluate models such as depth estimation~\cite{https://doi.org/10.48550/arxiv.2302.12288, depth_anything_v2, depth_anything_v1, Ganj_2025_WACV}, lighting estimation~\cite{xihe_mobisys2021, wang2022stylelight, Phongthawee2023DiffusionLight}, tracking\cite{zhang2015visual}, etc. 
\sysname allows the user to place the virtual object with different configurations, e.g., world position, distance to the camera, lighting conditions, and scales.

By adjusting these parameters, users can quickly assess how different model choices affect visual alignment and realism. 
Fig.~\ref{fig:depth_casestudy_screenshot} shows a screenshot of \sysname UI of object placement and rendering task of a captured scene with three depth estimation models (original, DepthAnything, and Zoedepth).

\para{Occlusion Rendering.} To ensure correct placement and interaction with the real world, it's important that occlusions are handled accurately. This task also gives us insight into the rendering quality of objects. For example, if the depth model fails to produce detailed depth maps, artifacts often appear around the rendered objects.
To better evaluate this, we have a black virtual plane that can be placed at various depth levels to observe how well the model captures edges and fine details. This can also serve as a useful tool for assessing the metric accuracy of depth models, as reviewing the placement between overlapping objects can reveal how precise the predicted depth really is.

\para{3D Point Cloud.}
This task provides a 3D view of object placement within the environment and complements the other two tasks by allowing users to inspect the scene from different angles. The point cloud view gives an intuitive sense of the reconstructed geometry and makes it easier to spot artifacts such as surface warping, holes, or depth discontinuities that may be hard to notice in 2D views. As we show in \S\ref{sec:caseStudy}, this feature helps uncover insights about depth model accuracy. Rendering is offloaded to a server, which uses scene data, camera parameters, and model outputs to configure the pipeline for the AR tasks.

\para{AR Tasks on Datasets.}
In addition to supporting model evaluation on data captured directly using our data collection tool, \sysname also provides the same AR evaluation pipelines for commonly used datasets, such as ARKitScenes~\cite{arkitscenes} and ScanNet~\cite{dai2017scannet}. Just like the pluggable model support, researchers can also use their dataset in \sysname by implementing the corresponding data loaders. Automatic AR rendering with datasets to provide plausible scenes can itself be challenging, and we discuss the potential solutions in \S\ref{subsec:extensions}.

\vspace{-0.4mm}

\subsection{Automatic Scenario Generation \& Custom Experiment Protocols}
During the design of visual perception experiments, conducting studies is often valuable for assessing potential question bias and difficulty levels. Initial findings from these studies can help researchers identify visual materials that are likely to yield meaningful insights for the final experiment. \sysname enables researchers to easily collect new data, visually and quantitatively explore results from commonly used datasets, and through live streaming and replay of AR scenes. In the following sections, we describe two extensions that researchers can build on top of \sysname.

\para{Automatic High-Quality Scenario Generation.}
Generating AR scenarios for debugging and perception studies means placing virtual objects in realistic positions. Doing this by hand takes time and often produces scenes that look wrong, such as a teapot floating in the air instead of sitting on a table, which can distract the study from focusing on rendering quality.
To streamline this process, we design an automatic re-rendering pipeline that generates AR scenarios by sampling object poses on detected support surfaces, such as tables or floors, and discarding candidates that would be out of the camera frustum. Figure~\ref{fig:autoPlacement} shows virtual teapots that our system places at different positions in the same captured scene.

\para{Custom Experiment Protocols.}
\label{subsec:extensions}
Researchers can easily extend \sysname by implementing custom experiment protocols. For instance, to incorporate the HYPE protocol~\cite{Zhou2019-hd}, which evaluates time-limited perceptual thresholds and human error rates, researchers can record real-world data using \sysname or provide a data loader for their target datasets. They can then use the edge-based processing pipeline to generate the required visual assets and implement the application logic that controls \emph{how, when, and what} visual content is shown as participants interact with the study interface. Supporting such custom protocols reduces setup effort and promotes reproducible evaluation of visual CV models. In \S\ref{sec:caseStudy}, we show how the three-sphere protocol~\cite{wang2022stylelight}, a custom lighting estimation evaluation procedure, can be integrated into \sysname to facilitate robust assessment of lighting estimation models.

\begin{figure}[t]
    \centering
    \includegraphics[width=\linewidth]{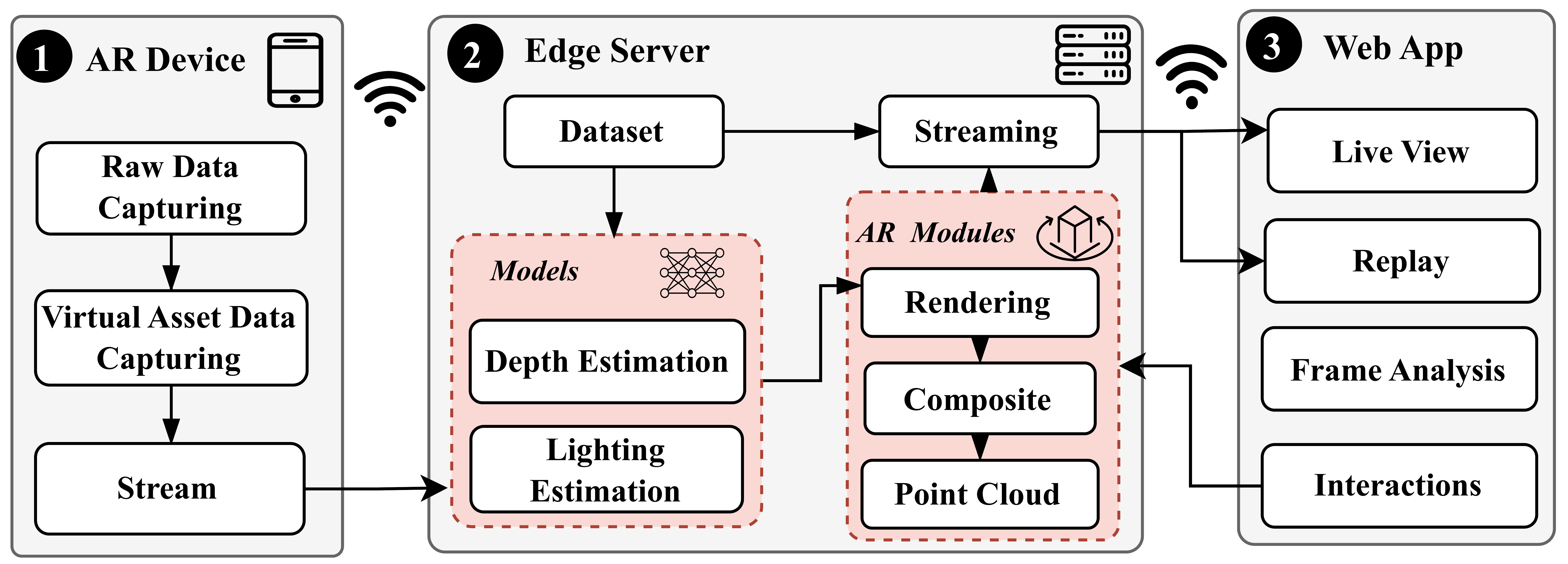}
    \caption{\sysname architecture.
    \textnormal{Illustration of its three main components and their associated modules.}
    }
    \label{fig:architecture}
    \vspace{-0.5cm}
\end{figure}

\begin{table*}[t]
\centering
\caption{
Component-based performance measurement across three resolutions. 
\textnormal{We use \textit{Teapot.obj} as the 3D asset and \textit{DepthAnythingV2-Base} as the pluggable CV model for measuring virtual object rendering and inference time. The AR device, server, and user web browsers are connected via the same university Wi-Fi. Results are averaged over 10 frames. Initialization costs range from 190 ms to 198 ms across all resolutions and are incurred only once.
}
}

\label{tab:component_perf}
\small
\begin{tabular}{l|c|c|c|c|c}
\toprule
\textbf{Resolution} & \textbf{Data Collection (ms)} & \textbf{Inference (ms)} & \textbf{Virtual Object Rendering (ms)} & \textbf{Composite (ms)} & \textbf{AR Streaming (ms)} \\
\midrule
640 $\times$ 480 (SD)     &  15.58 $\pm$ 7.2  & 45 $\pm$ 8  & 3.6 $\pm$ 1.6  & 1.6 $\pm$ 1.2   & 11.1 $\pm$ 1.2 \\
1280 $\times$ 720 (HD)    &  23.34 $\pm$ 8.1   & 69 $\pm$ 2  & 7.5 $\pm$ 0.6  & 4.1 $\pm$ 2.2 & 16.5 $\pm$ 2.3 \\
1920 $\times$ 1080 (FHD)  &  18.12 $\pm$ 9.5  & 74 $\pm$ 1  & 18.4 $\pm$ 0.35  & 5.3 $\pm$ 1.9   &  18.1$\pm$ 5 \\
\bottomrule
\end{tabular}%
\end{table*}

\subsection{Implementation}
\label{subsec:implementation}
\sysname is implemented as a microservice architecture with three main components: \circlednumber{1} an AR device for capturing and streaming data, \circlednumber{2} an edge server that runs models and AR modules, and \circlednumber{3} a web app for live view, replay, and interaction. These components run as independent services connected via an asynchronous Tornado backend; RGB, depth, and pose are streamed over WebSocket, while REST endpoints handle session control and model integration.

\para{\circlednumber{1} AR device.}
The AR client handles \emph{data capturing} and \emph{streaming}. We support a native iOS/ARKit app and a cross-platform ARFlow client~\cite{zhao2024arflow}. Both capture RGB ($1920{\times}1080$), depth (e.g., LiDAR depth at ARKit’s native resolution $192{\times}256$ from iPhone~14 Pro), and camera pose plus virtual-object metadata (mesh, pose, scale), and package each frame with intrinsics and nanosecond timestamps. RGB is PNG-compressed to preserve alpha channel, depth is sent as 16-bit buffers to avoid quantization, and metadata is JSON. Frames are streamed via WebSocket to the edge server; ARFlow can additionally use gRPC with gzip compression to reduce bandwidth. In our prototype, the device can be mounted on a small PiCar-X (Raspberry Pi 3B) for remote scene exploration, but the client can also run on other platforms (e.g., drones).

\para{\circlednumber{2} Edge server.}
The edge server hosts the \emph{dataset} store, \emph{models}, and \emph{AR modules}. Models are plugged in either as Python modules that follow a simple template or as Docker containers using a REST API. For this paper we implement depth and lighting estimation models and use them in our case studies (\S\ref{sec:caseStudy}). The server writes all frames and predictions to the dataset store and, for each frame, can compute standard metrics (e.g., RMSE, MSE, etc). AR modules implement \emph{rendering}, \emph{composite}, and \emph{point cloud} operations: virtual assets are rendered off-screen with PyRender or Three.js using the captured intrinsics, then composited on the RGB frame via a GPU-accelerated compositor. Point-cloud views are built with Open3D/Three.js. These captured sessions are stored, and the same physical scene can be automatically re-rendered with different models without re-capturing data.

\para{\circlednumber{3} Web app.}
The web app exposes the system to users through four functions: \emph{live view}, \emph{replay}, \emph{frame analysis}, and \emph{interactions}. It connects to the edge server over WebSocket and uses REST calls to switch and configure models and tasks, change AR tasks and interact with renderings. Researchers can watch live AR output, go through recorded sessions, compare different model predictions on the same frames, inspect per-frame metrics, and interact with AR tasks (e.g., object placement, plane selection, point-cloud navigation) directly in the browser. This makes it easy to move from numerical summaries to visual debugging and perception-based evaluation without leaving a single interface.

\vspace{-0.2cm}
\section{Performance Evaluation \& Case Studies}\label{sec:implementation_evaluation}

\begin{table}[t]
\centering
\caption{
Frame rate (FPS) for live streaming and replay streams at different resolutions over campus Wi-Fi. 
\textnormal{
Live streaming reflects real-time transmission from the AR device to the browser, while replay denotes the performance when rendering recorded sequences.
}
}
\label{tab:fps_network}
\small
\begin{tabular}{l|c|c}
\toprule
\textbf{Resolution} & \textbf{Live Streaming (FPS)} & \textbf{Replay (FPS)} \\
\midrule
640 $\times$ 480 (SD)     & 22 & 30 \\
1280 $\times$ 720 (HD)    & 14 & 27 \\
1920 $\times$ 1080 (FHD)  & 10 & 22 \\
\bottomrule
\end{tabular}
\vspace{-0.4cm}
\end{table}

\begin{table}[t]
\centering
\caption{
Interaction latency. 
\textnormal{Measurements are over an institute campus Wi-Fi network and represent the end-to-end responsiveness of each interaction type. RC Car Latency for sending and executing commands is 38 ms.
}
}
\label{tab:interaction_latency}
\small
\begin{tabular}{l|c|c}
\toprule
\textbf{Resolution}  & \textbf{Point Cloud (ms)} & \textbf{Obj Interaction (ms)} \\
\midrule
640 $\times$ 480 (SD)     & 19 $\pm$ 3.2  & 7.52 $\pm$ 2.5 \\
1280 $\times$ 720 (HD)    & 33.5 $\pm$ 7.3  & 12.3 $\pm$ 10.5 \\
1920 $\times$ 1080 (FHD)  & 57.1 $\pm$ 9  & 10.2 $\pm$ 5.6 \\
\bottomrule
\end{tabular}
\vspace{-0.5cm}
\end{table}

We evaluate \sysname from two dimensions: system performance/usability and usefulness via two case studies.

\para{Configuration.} We configure \sysname with a unified setup where captures are collected on an iPhone~14 Pro and streamed to an edge server equipped with an NVIDIA RTX~4090 GPU, a 13th~Gen Intel i9-13900K CPU, and 128GB of RAM. All inference is executed on the server, and both benchmark metrics and visual outputs are recorded for replay. Experiments are conducted on campus Wi-Fi where the AR device, edge server, and web client share the same access point.

\para{Dataset.}
To support our case studies, we built a small evaluation dataset. 
All sequences were recorded with an iPhone~14 Pro, streamed to the edge server, and saved for replay across different tasks and models. 
The dataset includes a 407-frame video, designed to highlight challenges such as reflective surfaces, large occlusion boundaries, and complex indoor lighting. Although the dataset is not large, it serves as a controlled benchmark for \sysname: each capture can be reused across multiple evaluation pipelines, enabling fair comparisons and reproducible case studies.

\vspace{-0.4cm}
\subsection{Performance Evaluation}
\label{Evaluation}

\para{Micro-benchmark.}
Table~\ref{tab:component_perf} details the per-component latencies of our system across three different resolutions. Due to the optimizations and implementation choices that we discussed in \S\ref{subsec:implementation}, our current configuration achieves real-time performance. Notably, based on the optimizations that we have for streaming and client-side compression, our optimized rendering and compositing pipeline reduces latency significantly from an initial baseline of approximately 300\,ms per frame to just 5.2\,ms at 640$\times$480 resolution. 

\para{Interactions.}
Table~\ref{tab:interaction_latency} reports end‑to‑end latencies for two interactive features—object/plane manipulation and RC‑car control. We observe that object and plane updates are highly responsive. The object interaction latency ranges from 7.5~ms at SD to 12~ms at FHD, so adjustments appear almost instantaneously. RC‑car commands average 38~ms round‑trip over the same Wi‑Fi network, and execution on the vehicle side adds only about 1~ms.

\begin{figure*}[t]
    \centering
    \begin{subfigure}[t]{0.45\linewidth}
        \centering
        \includegraphics[width=\linewidth]{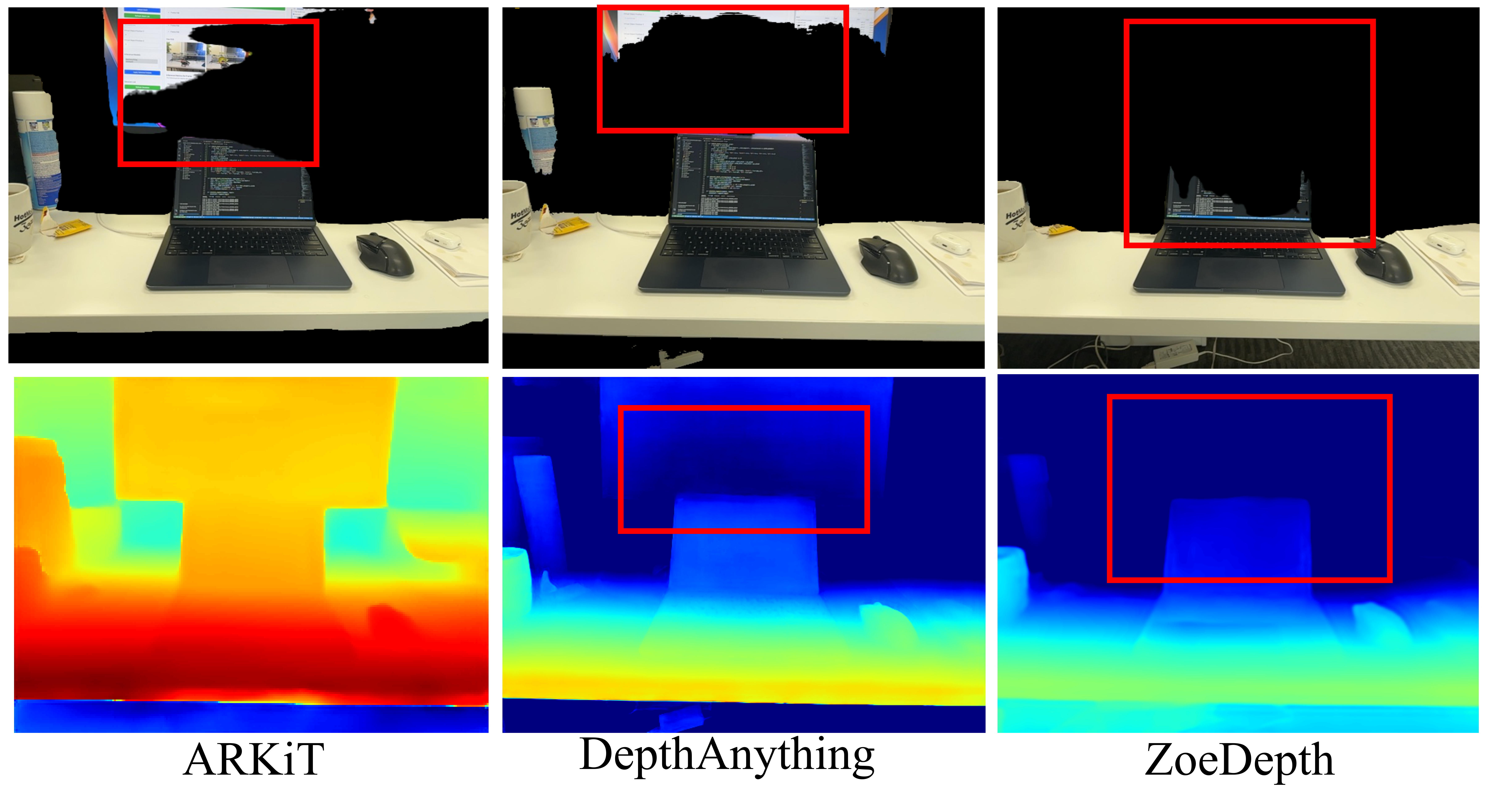}
        \caption{
        Plane: 0.75 m for ARKit; 1.3 m for DepthAnything/ZoeDepth. 
        \textnormal{Visualization highlights depth estimation differences across methods.}}
        \label{fig:depth_cs_dynamic_plane}
    \end{subfigure}
    \hfill
    \begin{subfigure}[t]{0.45\linewidth}
        \centering
        \includegraphics[width=\linewidth]{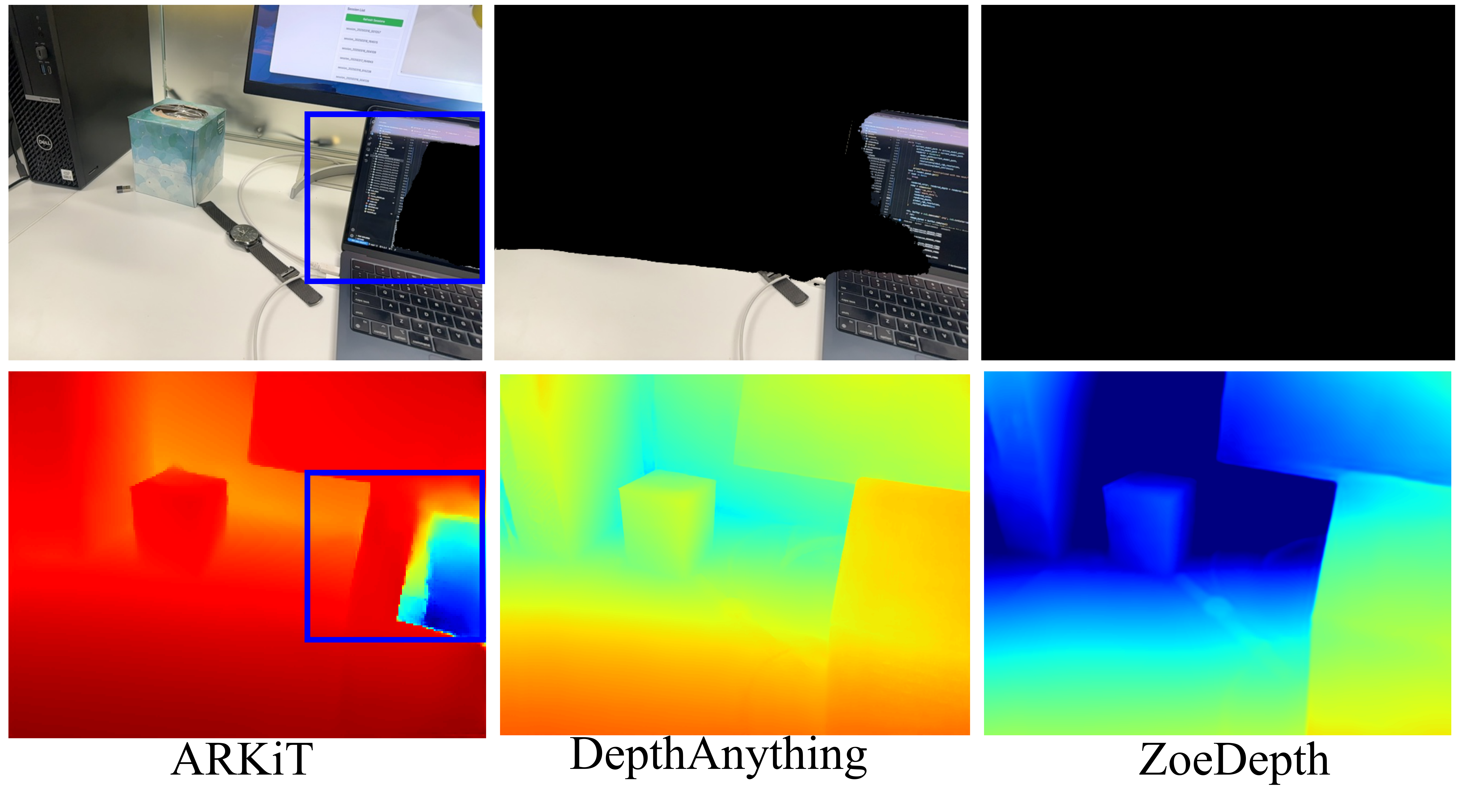}
        \caption{Plane set to 0.95 m. 
        \textnormal{
        Different frames of the same scene are shown to evaluate spatial detail and temporal smoothness.
        }
        }
        \label{fig:depth_cs_fixed_plane}
    \end{subfigure}
     \vspace{-0.2cm}
    \caption{Depth visualization at different plane distances and across different frames of the same scene. \textnormal{Red represents closer depth values, while blue indicates farther distances.}}
    \label{fig:depth_casestudy_plane}
    \vspace{-0.3cm}
\end{figure*}

\para{Replay and Live Stream.}
Table~\ref{tab:fps_network} shows the frame rates achieved in both live‑streaming and replay modes. Replay meets real‑time requirements at every resolution because frames are rendered on the server from pre‑recorded data, avoiding network upload and on‑device delays. Live streaming reaches 22~FPS at $640\times480$, which satisfies the frame‑rate target for most mobile AR applications.
For point cloud streaming, we transmit data in the \texttt{.pcd} format, which is roughly 75\% smaller than \texttt{.ply}. Combined with Open3D’s built‑in compression, this keeps point cloud latency below 60~ms even at 1080p. 
We found this latency sufficient for smooth 3D scene analysis in our experiments.


\para{Cross-Renderer Geometry Correctness.}
To ensure that server-side evaluation matches on-device behavior, we verify that our renderer faithfully reproduces ARKit geometry. We snapshot the virtual layer on the phone in a parity mode (solid background, unlit material, no shadows) and render the same objects on the server with matched intrinsics and poses. We then compare binary masks using Silhouette IoU, Boundary F-score (2\,px tolerance), and scale error, isolating geometry from photometric effects. Across tested scenes, Silhouette IoU and Boundary F1 are $1.0 \pm 0.0$ and scale error is $0.0 \pm 0.0$\%, indicating that server-side rendering is geometrically identical to ARKit while remaining scriptable and scalable.

\subsection{Case Studies: Depth and Lighting}
\label{sec:caseStudy}

Depth estimation and lighting estimation are long-standing CV problems with established datasets and metrics, such as RMSE or $\delta$ thresholds for depth, angular error or RMSE for lighting.
However, numerical evaluations often fail to capture how models behave in realistic visual contexts.
Small numerical differences can correspond to large perceptual artifacts, and in some cases different metrics even lead to contradictory rankings. 
To highlight these limitations, we conduct two case studies using \sysname: depth estimation as an example of a geometry-centric CV task, and lighting estimation as an appearance-centric task. Together they illustrate two points: first, the usefulness of \sysname in exposing evaluation blind spots through perceptual visualization; and second, the findings that such analysis reveals about state-of-the-art CV models.

\subsubsection{Depth Estimation}
We use three representative depth pipelines: Depth Anything V2 (metric depth version)~\cite{depth_anything_v2}, ZoeDepth~\cite{https://doi.org/10.48550/arxiv.2302.12288}, and ARKit~\cite{AppleARkit}. 
\sysname supports both raw depth-map visualizations and interactive plane placements to evaluate occlusion consistency across different models. In Fig.~\ref{fig:depth_casestudy_plane}, we place virtual planes in each scene to assess how accurately and consistently each model predicts depth.
Below we list key findings using \sysname.

First, we observe significant variation in global scale across methods. In Fig.~\ref{fig:depth_cs_dynamic_plane}, the virtual plane needs to be positioned at 0.75 m when using ARKit depth. However, to achieve an equivalent occlusion with DepthAnythingV2 and ZoeDepth, the plane must be placed nearly twice as far, at 1.3 m. This reveals a consistent overestimation of metric depth by these models, which can severely impact spatial coherence in AR scenes. In Fig.~\ref{fig:depth_cs_fixed_plane}, we fix the plane at 0.95~m across all methods. Ideally, the occlusion should appear similar in each view, but significant discrepancies remain, highlighting issues with depth scale consistency.

Second, the quality of local boundaries varies notably. For example, ARKit struggles with reflective surfaces like a laptop monitor, shown in the blue boxes (Fig.~\ref{fig:depth_cs_fixed_plane}), causing the occlusion plane to appear incorrectly over the monitor. Similarly, ZoeDepth completely fails to capture the structure of the large monitor, as highlighted by the red box in Fig.~\ref{fig:depth_cs_fixed_plane}. While these artifacts may be subtle in the color-coded depth maps, they become starkly visible in the rendered occlusion views, emphasizing the value of task-driven visualizations.

\begin{figure}[t]
    \centering
    \includegraphics[width=\linewidth]{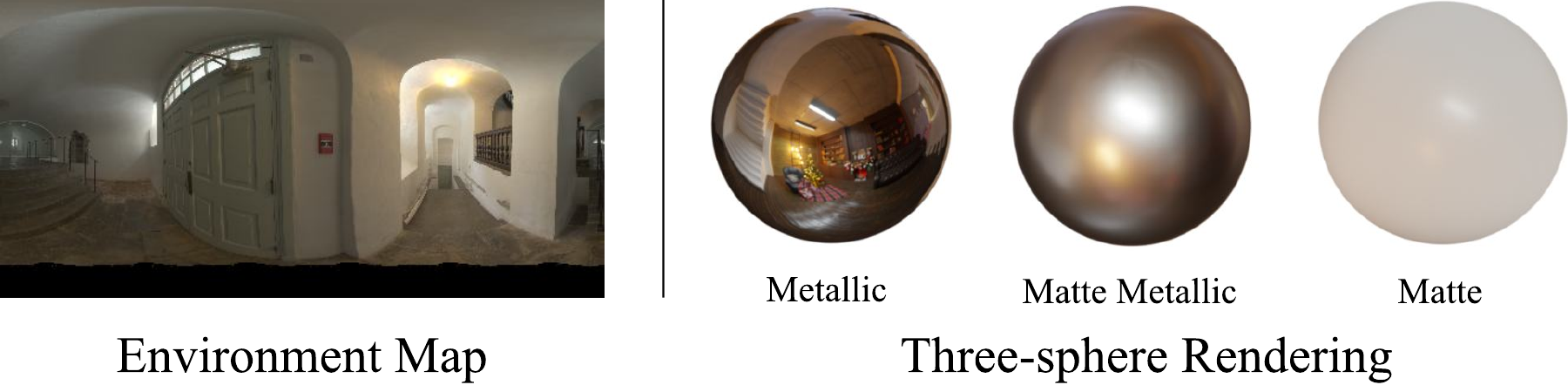}
    \caption{Lighting estimation experiment protocol overview. \textnormal{We evaluate the accuracy of lighting estimation models by directly comparing the estimated environment maps (left) and three rendered virtual spheres of different materials (right).}}
\label{fig:experiment_protocols}
\vspace{-0.3cm}
\end{figure}


\begin{figure}[t]
    \centering
    \includegraphics[width=0.95\linewidth]{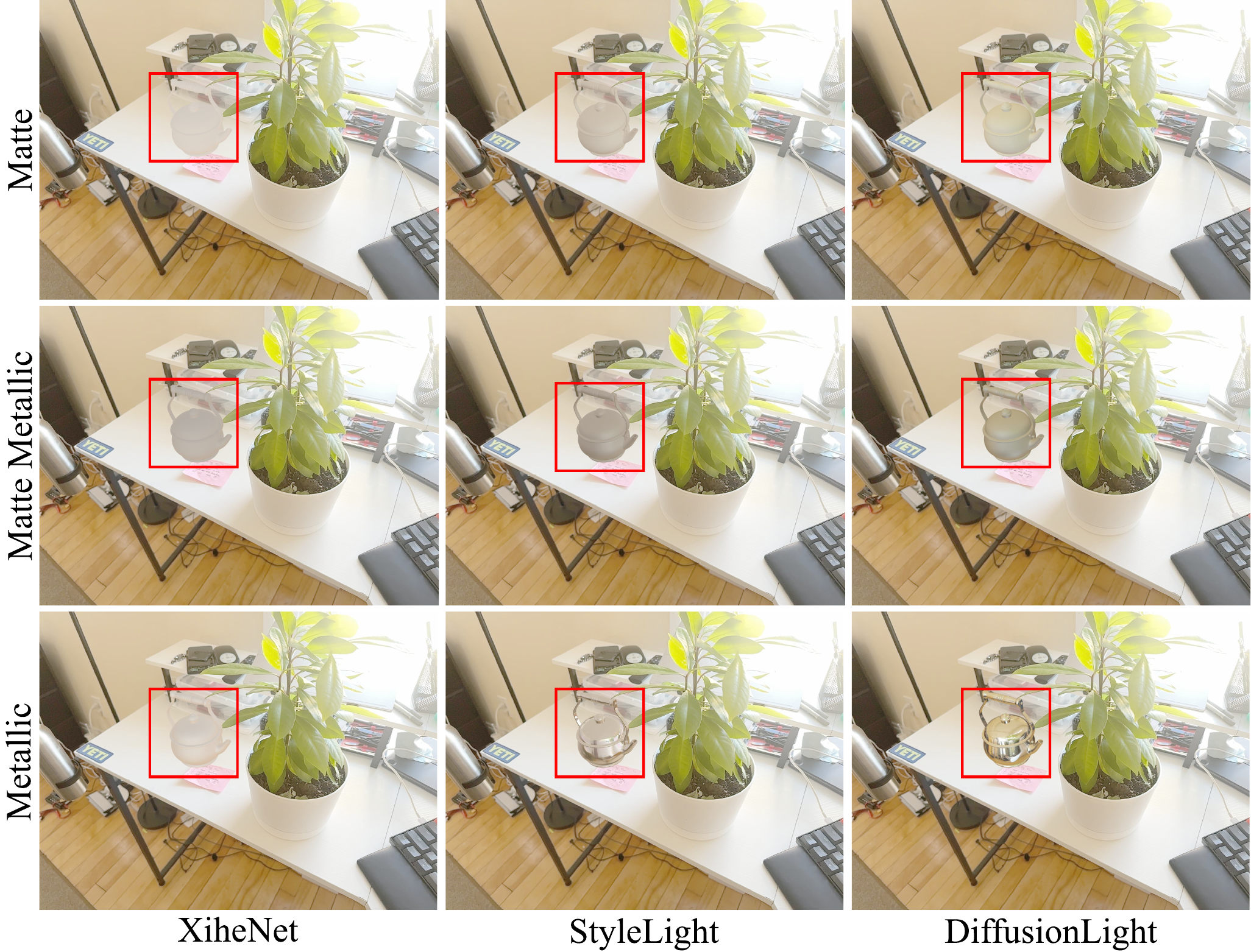}
    \caption{Lighting estimation rendering visualizations. \textnormal{The \sysname visualization interface enables side-by-side display of virtual teapots under varying rendering parameters, such as different materials and lighting conditions. This design facilitates perception-centered evaluation studies by supporting frame-by-frame visualization of numerical results from Table~\ref{tab:eval_three_sphere_reorganized}}.
    }
\label{fig:case_study_lighting_result_visualizations}
\vspace{-0.5cm}
\end{figure}

\subsubsection{Lighting Estimation}

We evaluate three representative lighting estimation models—XiheNet~\cite{xihe_mobisys2021}, StyleLight~\cite{wang2022stylelight}, and DiffusionLight~\cite{Phongthawee2023DiffusionLight}; spanning lightweight, mid-sized, and large architectures. All three are plugged into \sysname as containerized models and evaluated under a common protocol. For numerical evaluation, we use the Laval Indoor HDR dataset~\cite{Gardner2017} and report angular error, Si-RMSE, and RMSE. For task-level evaluation, we implement the \emph{three-sphere protocol}~\cite{wang2022stylelight} as a built-in experiment: Metallic, matte Metallic, and Matte spheres are rendered in AR under the estimated lighting using our standard rendering pipeline (Fig.~\ref{fig:experiment_protocols}).

As discussed in \S\ref{sec:perceptionGap}, metrics do not always agree; different numerical metrics often highlight different aspects of lighting estimation performance. To better understand these numerical discrepancies, we use \sysname to visualize each model’s predictions under the same input condition. As shown in Fig.~\ref{fig:case_study_lighting_result_visualizations}, relit teapots rendered using different models reveal perceptual differences not captured in the numerical scores. While quantitative scores rank DiffusionLight highest, side-by-side relighting reveals artifacts such as overly bright teapots or missing directional shadows; XiheNet, in contrast, tends to oversmooth specular highlights, making metals appear dull. In contrast, DiffusionLight sometimes overestimated intensity, causing teapots to look unnaturally bright. These nuanced, material-dependent failures are hard to infer from environment maps or metrics, but they become obvious when rendered into the AR scene using \sysname's pipeline. This case study illustrates how \sysname's design connects numerical lighting metrics to the perceptual quality that matters in AR applications.

\vspace{-0.1mm}

\subsubsection{Discussion and Implications}

Our case studies highlight the central message: benchmark numbers alone do not guarantee perceptual quality in the real world. DepthAnything V2 and ZoeDepth, despite strong performance on NYU Depth V2, still overestimate scale and blur object boundaries, leading to artifacts in placement and occlusion. Lighting models such as XiheNet, StyleLight, and DiffusionLight show a similar gap: two models with comparable angular error or RMSE can produce noticeably different renderings, especially on reflective or complex materials. By reusing the same captured scenes across all model–metric–task combinations, \sysname enabled us to diagnose these issues under identical conditions, combining metrics with direct AR visualizations such as point clouds, occlusion renders, and relit objects. This capture-once-evaluate-many workflow revealed temporal instabilities, scale errors, and lighting artifacts that benchmarks alone obscure, showing how \sysname complements traditional metrics and leads to a more complete understanding of the CV model.

\section{User Study}

We conducted an IRB-approved user study with fifteen participants, all of whom have prior experience working with CV. The study examines two main questions: what challenges researchers see in CV evaluation and how effective \sysname is in addressing them.

\para{Study Protocol.} The study had three parts. First, participants completed a short Qualtrics survey on their role, research area, CV/ML experience, typical evaluation workflow, and attitudes toward dataset-centric metrics versus application-driven evaluation; questions were modeled after the challenges in \S\ref{sec:challenges}. Second, they received a brief introduction to \sysname and a guided tutorial covering its core features, including live AR streaming, replay of recorded sequences, virtual object rendering, occlusion rendering based on depth models, 3D point cloud visualization, and basic virtual object interactions. Third, participants used \sysname to capture an AR scene on their own and then completed a post-study survey with five-point Likert ratings on overall satisfaction, perceived usefulness for judging task suitability and identifying failure cases, and feature-level usefulness, ease of use, and responsiveness.

\vspace{-0.3cm}
\subsection{Results \& Analysis}
\subsubsection{Background Study}
\label{subsec:background_study}
We received responses from 15 participants: 73\% PhD students and 27\% undergraduate students, all working on CV/ML-related topics. Their experience ranged from less than <1 to >4 years, providing a mix of new and more experienced practitioners. 
Participants reported high programming comfort (avg. 4.33/5) but more limited mobile-development skills (avg. 2.53/5).

Only 47\% of participants reported testing models in real-world scenarios (e.g., AR or robotics). Among those who do, \emph{engineering effort} was rated moderately high (mean 3.57/5); among those who do not, it was the top barrier. In a multi-select question, participants most often evaluated via public datasets/leaderboards (60\%) or private datasets (40\%), with fewer using application tasks (33\%) or user studies (27\%). These results suggest dataset-centric evaluation dominates, driven in part by the perceived cost of in-the-wild testing; \emph{\sysname\ can reduce this cost by lowering engineering effort.}

Participant responses also reinforce the limits of purely data-driven benchmarking and align with the challenges in \S\ref{sec:challenges} (Table~\ref{tab:bg-to-challenges}). Participants reported only moderate confidence that dataset metrics predict real-world performance (mean 3.0/5) and tended to disagree that common metrics are sufficient for most decisions (mean 2.67/5) or that leaderboard rankings transfer well to real applications (mean 2.60/5). In contrast, they strongly agreed that clear visualizations are as important as numeric scores (mean 4.53/5), that evaluation is sensitive to protocol choices such as splits and normalization (mean 4.13/5), and that sensors and hardware during data collection can introduce noise that reduces evaluation reliability (mean 4.47/5).

\begin{table}[t]
\centering
\caption{Survey findings that support the challenges in \S\ref{sec:challenges}.}
\label{tab:bg-to-challenges}
\small
  \resizebox{0.9\columnwidth}{!}{%
\begin{tabular}{@{}p{0.36\linewidth}p{0.6\linewidth}@{}}
\toprule
\textbf{Challenge} & \textbf{Key survey evidence} \\
\midrule
Limitations of Data-Driven Benchmarking (\S\ref{subsec:evaluation_limitations}) 
& Metrics are not sufficient for evaluation; leaderboard performance doesn't transfer\\[0.3em]

Data \& Labels (\S\ref{subsec:data_labels}) 
& Sensors/hardware introduce noise that affects the evaluation.\\[0.3em]

The Metric-Perception Gap (\S\ref{sec:perceptionGap}) 
& Clear visualizations are important; evaluation is sensitive to protocol choices. \\
\bottomrule
\end{tabular}
}
\vspace{-0.4cm}
\end{table}

\subsubsection{User Experience with \sysname}

\begin{figure}[t]
  \centering
  \includegraphics[width=0.9\linewidth]{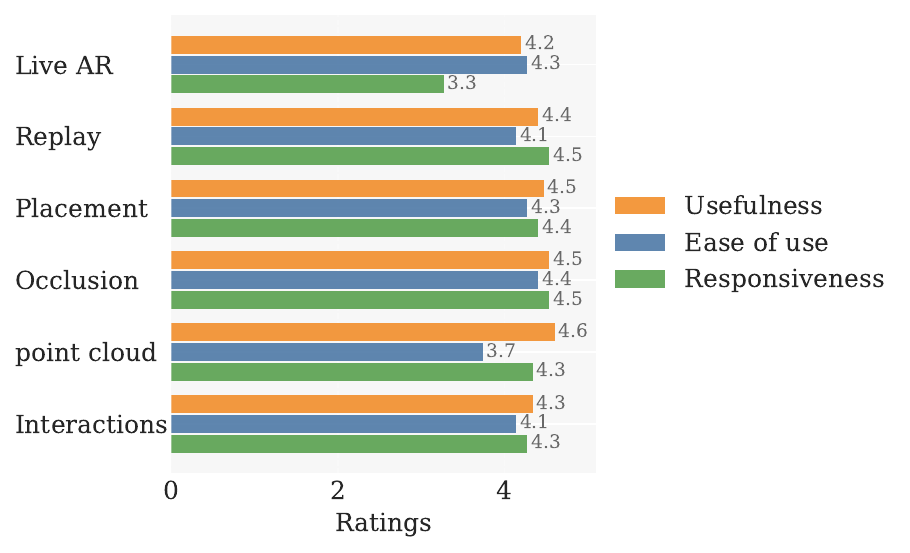}
  \caption{User Study ratings for each feature.}
  \label{fig:arcade-feature-ratings}
    \vspace{-0.6cm}
\end{figure}

After the hands-on session with \sysname, participants rated the overall experience and specific features. Overall satisfaction and intention to use a platform like \sysname in the future were both high, each with an average of 4.20/5. Participants judged \sysname to be very effective for supporting task-level decisions about depth models (avg. 4.67) and for helping them discover failure cases more easily (avg. 4.53). Usability ratings were positive: they found the system easy to learn (avg. 4.20) and generally easy to use in practice (avg. 3.93). They also agreed that \sysname reduces the engineering effort required for real-world evaluation (avg. 4.33), supporting \sysname's goal.

Figure~\ref{fig:arcade-feature-ratings} shows feature-level ratings for each major component of \sysname. Usefulness scores were consistently high across all features (at least 4.20), showing that \sysname's design choices are valid. Ease-of-use scores were generally high (between 4.1 and 4.4), with the 3D point cloud interface slightly lower at 3.73. This indicates the need to simplify it. Responsiveness ratings were highest for session replay and occlusion rendering (around 4.5). However, live AR streaming was lower at 3.27, underscoring streaming latency and stability as important future optimization.

\section{Conclusion \& Future Work}
\label{sec:conclusion}

We presented \sysname, an evaluation framework that bridges the gap between insufficient quantitative benchmarks and hard-to-do visual evaluation. By providing a reusable pipeline and interactive AR tasks, \sysname enables researchers to complement metrics with direct visual inspection. Our user study with early-stage researchers and case studies using \sysname demonstrated its usefulness in revealing significant perceptual flaws in SoTA that benchmarks missed. 
We envision that \sysname can be deployed by CV researchers in their labs with minimal setups and therefore lower the barrier to conduct principled metric and visual evaluations using end-to-end AR application pipelines. 
Looking ahead, we plan to broaden supported tasks and explore remote interaction extensions (e.g., robotic manipulators~\cite{Tung2021-uj} or virtual device streaming) so that off-site participants can interact with captured environments.

\begin{acks}
This work was supported in part by NSF Grants \#2236987 and \#2346133. 
\end{acks}

\balance
\bibliography{ref}

\end{document}